# A Localization and Navigation Method for an In-pipe Robot in Water Distribution System through Wireless Control towards Long-Distance Inspection

**Saber Kazeminasab[1], Student Member, IEEE, M. Kathrine Banks[2]**
[1]Department of Electrical and Computer Engineering, Texas A&M University, TX 77847 USA
[2]College of Engineering, Texas A&M University, TX 77847 USA

Corresponding author: Saber Kazeminasab (e-mail: skazeminasab@tamu.edu).

**ABSTRACT** In this paper, we propose an operation procedure for our previously developed in-pipe robotic system that is used for water quality monitoring in water distribution systems (WDS). The proposed operation procedure synchronizes a developed wireless communication system that is suitable for harsh environments of soil, water, and rock with a multi-phase control algorithm. The new "wireless control algorithm" facilitates "smart navigation" and "near real-time wireless data transmission" during operation for our in-pipe robot in WDS. The smart navigation enables the robot to pass through different configurations of the pipeline with long inspection capability with a battery in which is mounted on the robot. To this end, we have divided the operation procedure into five steps that assign a specific motion control phase and wireless communication task to the robot. We describe each step and the algorithm associated with that step in this paper. The proposed robotic system defines the configuration type in each pipeline with the pre-programmed pipeline map that is given to the robot before the operation and the wireless communication system. The wireless communication system includes some relay nodes that perform bi-directional communication in the operation procedure. The developed wireless robotic system along with operation procedure facilitates localization and navigation for the robot toward long distance inspection in WDS.

**INDEX TERMS** In-pipe Robots, Wireless Control Algorithm, Smart Navigation, Water Quality Monitoring, Water Distribution Systems.

## I. INTRODUCTION

Water Distribution Systems (WDS) are responsible to carry potable water to residential areas. Aging pipelines cause leaks and water loss in the system. The amount of water loss is unavoidable; around 15%-25% and 20% of the purified water is reported for water loss in the US and Canada, respectively [1], [2], Hence, it is required to periodically assess the condition of pipelines and localize the leak location. The traditional methods for leak detection primarily depend on user experience [3]. In addition to condition assessment and leak detection, utility managers need to measure water parameters periodically to ensure the health of water. However, it is an extremely challenging task to access all parts of the distribution network as the pipelines are long, composed of complicated configurations, and commonly buried beneath the earth's surface. Mobile sensors that are designed to go inside pipes are not feasible to use as their motion depends on flow in the pipe and the user loses them in the network during operation [4], [5]. In-pipe robots are designed to solve the problem of passiveness of mobile sensors, in which actuator units and control algorithms control their motion to be independent of water flow. They move inside pipes and perform desired tasks (e.g. leak detection or quality monitoring) [6] and are powered by either cable [7] or battery [8]. Due to the long length of pipelines and the short length of cables, the feasibility of tethered robots are limited. As a result, battery-powered robots are desirable for long pipeline inspection. The battery-powered in-pipe robots communicate with the base station above ground through wireless communication [9]. However, due to high path loss and dynamic communication links of soil, rock, and water, wireless communication is a challenging task in these environments [10], [11]. This makes the range of coverage of current communication setups limited and it is not possible to cover all parts of the network with one pair of transceivers. To address the problem of limited coverage, the idea of relay nodes is introduced in the literature [12] in which some transceivers are placed above ground and relay the data between robotic sensors and base station. Relay nodes are usually used for fixed sensor networks [13]. A team from the

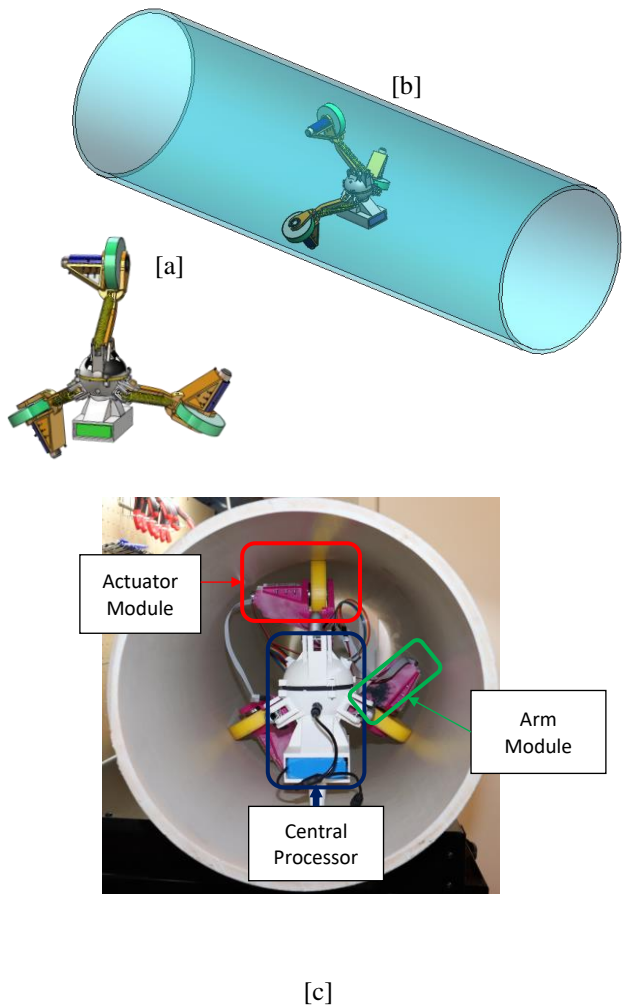

**Figure 1.** Our Proposed Self-powered In-pipe Robot: [a] CAD Design. [b] Robot in Pipe. [c] Prototype of the Robot in a Pipe and the Components: Actuator Module, Arm Module, and Central Processor.

Massachusetts Institute of Technology (MIT) has used the idea of relay nodes in in-pipe robotic sensors and coined the term "wireless robotic sensors" for localization of the robot [14]. Wireless robotic sensors are advantageous over other methods as longer distances of the network can be inspected while the sensor measurements can be transmitted during operation.

### A. TECHNICAL GAP

Pipelines are composed of different configurations like bends and T-junctions in which the robot needs to pass through them to facilitate long pipe inspection. To the best of our knowledge, there is not a method to localization and navigation in chattered and complicated environment of in-service pipelines with long distance inspection. Also, it is desired the in-pipe robot perform regular measurement of water parameters (e.g. turbidity, conductivity) or pipe parameters (e.g. pressure gradient) during operation that is not well-addressed in the literature.

### B. OUR CONTRIBUTION

In this research:
1) We improve our previously designed in-pipe robot [9] for locating the electronic parts indie the central processor.
2) We propose a bidirectional low-frequency wireless system and synchronize it the micro-pump system and motion control unit.
3) We propose a localization method to localize the robot in complicated configurations of pipelines.
4) We propose a navigation method to enable the robot move in complicated configurations of the pipeline.
5) We develop an operation procedure for the robotic system that facilitate regular measurements of parameter in water in long distance inspection.

## II. PROBLEM STATEMENT

Current in-pipe robots lack smart motion during operation in pipelines. To facilitate long inspection and wireless communication for in-pipe robots, we use a robotic system that is equipped with low-frequency wireless communication system. The robot moves inside pipeline and some wireless nodes are placed above ground. These nodes are placed at non-straight configurations and guide the robot to the desired path in the pipeline.

The remainder of the paper is organized as follows: In Section II, the design of the modular in-pipe robot is present, modeled, and characterized. In Section III, the wireless sensor module is developed and validated. In section IV, the operation procedure for the robotic wireless is presented and the verified with the experimental results, and in Section VI, the paper is concluded.

## III. ROBOT DESIGN, MODELING, AND CHARACTERIZATION

### A. DESIGN

The moving part of the robotic sensor network is a modular robot that is located inside pipelines. This robot is composed of:
- A Central Processor.
- Three Arm Modules.
- Three Actuator Modules.

Three arm modules are connected to the central processor at 120° angles and make the outer diameter of the robot adaptable to the pipe diameter. The design of the arms is similar and they rotate around the central processor without friction by ball bearings. A passive spring is located between the arm and the central processor in which one end is anchored on the arm and the other end is anchored on the central process or. The passive spring provides the press force for the wheel and also makes the diameter of the arm adjustable to pipe diameter (9 in-22 in).

The robot moves inside the pipe by actuator modules that are connected at the end of the arm modules. Each actuator module comprises a gear motor, a motor cover, moldable glue, and a wheel. The gear motor is fastened to the arm by the motor cover and moldable glue. Also, it connects to a wheel directly in which the wheel contacts and presses the pipe wall. The wheel is located on the arm and rotates friction-free that is facilitated by a pair of sealed ball bearings. More detail on the design, characterization, and fabrication can be found in our previous work [15]. The central processor is composed of two hemisphere-like parts that are fastened together with an O-ring and four pairs of screw-nuts. The central part is designed to locate the sensing, control, drive, wireless, and power units of the robot. The robot is self-powered and the battery is located below the control part and its power is transferred to other electronic units via wire. The electronic units (i.e. sensing, control, wireless, and drive units) are designed and printed on a printed circuit board (PCB). A micro-pump system provides water samples for the sensors in the central processor through water inlets and outlets and its function (i.e. timing) is controlled with the micro-pump driver. Fig. 2 shows the improved central processor, the PCB and its location the central processor, the way the wires from the motors and motors' encoder wires are directed and connected to the PCB, water inlets and outlet, and wireless system antenna. We will explain about the wireless system later in this paper.

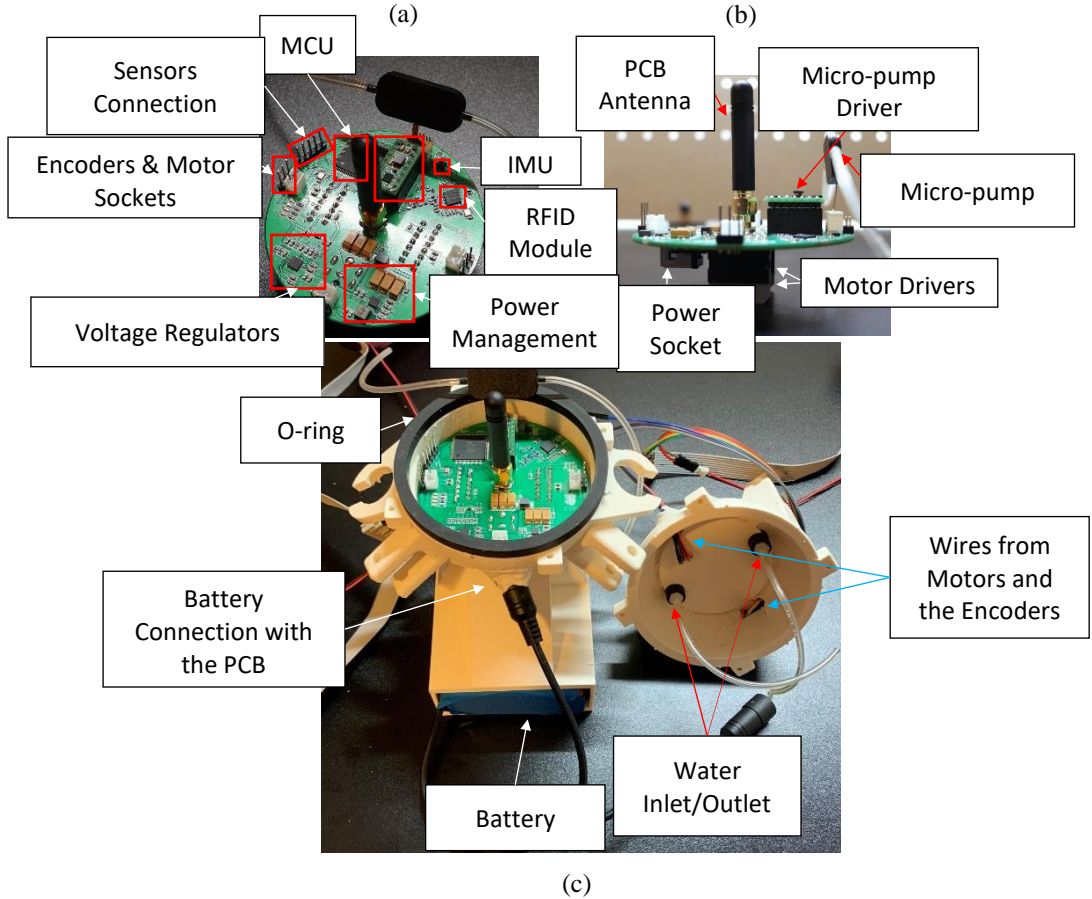

**Figure 2.** PCB and Central Processor. (a) Perspective View of the PCB. The sensors, encoders and the motors wires are connected to the PCB via headers. Three voltage levels of 12V, 5V, and 3.3V is available in the PCB via voltage regulators. (b) Side View of the PCB. Micro-pump system provides water sample to the water sensors for parameter measurements (e.g. PH, free-chlorine) and the timing for micro-pump is controlled via the MCU and the micro-pump driver. (c) PCB Location in the Central Processor and its Connection with the Battery. The wireless from the motors and the Water inlets and outlets are defined in the figure.

### B. MODELING
#### 1) ROBOT:

The robot design is similar to the two-wheeled inverted pendulum robots [16]–[24] with the difference that it has three wheels and the pendulum needs to be kept at a horizontal position. In Fig. 3, we can assume that the central processor has negligible linear displacement in the y and z axes and rotation in the x-axis. Hence, the dynamic of the robot can be presented with the following equations [25]:

$$\ddot{x} = f_1(V_f, \dot{x}, \boldsymbol{u}) \quad (1)$$
$$\ddot{\phi} = f_2(\phi, \psi, \theta_1, \theta_2, \theta_3, \boldsymbol{u}) \quad (2)$$
$$\ddot{\psi} = f_3(\phi, \psi, \theta_1, \theta_2, \theta_3, \boldsymbol{u}) \quad (3)$$

In (1)-(3), $\boldsymbol{u}$ is the input vector to the system in which $\boldsymbol{u} = \begin{bmatrix} \tau_1 \\ \tau_2 \\ \tau_3 \end{bmatrix}$. $\tau_1$, $\tau_2$, and $\tau_3$ are motor torques. We will provide modeling for gear motors in Section III-B2 to calculate the torques.

#### 2) GEAR-MOTOR

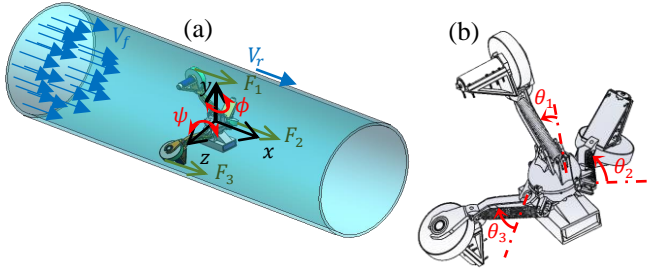

**Figure 3.** Forces Applied on the Robot During Operation. (a) Robot in Pipe. (b) Definition of Arm Angles.

The robot has three gear motors that are directly connected to wheels. Dynamical behavior of the gear motors are written as:

$$\frac{v_{co}}{L_m} - \frac{v_e}{L_m} - \frac{R_m}{L_m} i_m = \frac{di_m}{dt} \quad (4)$$

In (4), $R_m$, $L_m$, and $i_m$ are terminal resistance, terminal inductance, and the current through the gear motor, respectively. $v_{co}$ is the input voltage (control signal) and $v_e$ is the back EMF voltage and $v_e = k_v \omega$ ($k_v$ is motor characteristic and constant, and $\omega$ is the shaft angular velocity). $\tau_m = k_v i_m$ in which $i = 1, 2, 3$.

### C. Characterization

There is a notion of resilient machine that accounts for the requirements to design a system that is robust against the undesired disturbances and uncertainties [26]. In this regard, we defined the condition where the robot fails during operation and improve the design to prevent failure [9]. The robot s supposed to operate in in-service networks that are pressurized and water flow is present. Hence it is important that the motion of the robot is independent of flow motion. In our robot, two components paly important role in the design of automotive system and we characterize them in this section.

#### 1) SPRING MECHANISM

Since the robot motion is based on wheel wall-press, friction force between the pipe wall and the wheels needs to be sufficient to have pure rolling and prevent slippage. We need to provide an analysis for the friction to prevent slippage of the wheel to have controlled motion. The friction is provided by the passive spring attached on each wheel ($F_N$ and $F'_N$ in Fig. 4a). Since the weight of the robot increases the normal force of the wheels that are below center of mass of the robot (Fig. 4a), the required stiffness for the springs that are attached to these arms and wheels can be less than the required stiffness for the wheel that is above the center of mass (Fig. 4a). Hence, we provide the analysis for this wheel that is more likely to lose its contact with the pipe wall during motion. Based on Fig. 4b, we can write:

$$\sum M_O = 0 \rightarrow (mg - F'_N) a \cos\beta = f_s \cdot H + F_{Spring} \chi_{Spring} \quad (4)$$

In triangle OAB:

$$\beta = \alpha + \left(\frac{\pi}{2} - \theta\right) \quad (5)$$

$$\alpha = \sin^{-1}\left(\frac{t}{a}\cos\theta\right) \quad (6)$$

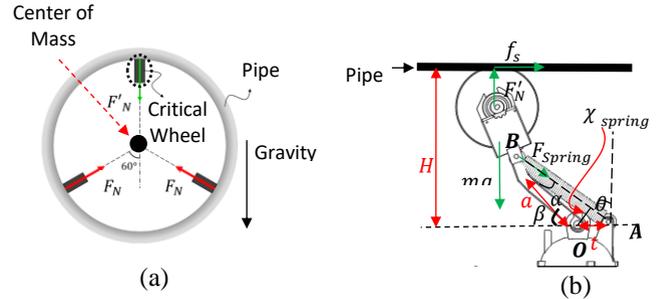

**Figure 4.** (a) The Critical Wheel. (b) Free Body Diagram of the Critical Wheel for Static Force Analysis.

From (5) and (6), we have:

$$\beta = -\theta + \sin^{-1}\left(\frac{t}{a}\cos\theta\right) + \frac{\pi}{2} \quad (7)$$

Also, we can write:

$$\beta = \sin^{-1}\left(\frac{H}{L}\right) \quad (8)$$

Plugging (8) to (7):

$$\sin^{-1}\left(\frac{H}{L}\right) = -\theta + \sin^{-1}\left(\frac{t}{a}\cos\theta\right) + \frac{\pi}{2} \quad (9)$$

In (9), $\theta$ is calculated with a nonlinear trigonometric relation based on pipe radius, $H$. We also have:

$$\chi_{Spring} = t\cos\theta \quad (10)$$

Plugging (10), (9), and (7) into (4):

$$F_{Spring} = \frac{1}{t\cos\theta}\left((F'_N - mg)a\cos\left(\theta + \sin^{-1}(\frac{t}{a}\cos\theta)\right) - f_s H\right)$$
(11)

The springs are linear where the relation between force, $F_{Spring}$, and displacement for them are linear. The displacement is calculated between the points where $\theta = 0$, $\sqrt{(t+\cos\beta)^2 + (a\sin\beta)^2}\cos\theta$ and $\theta > 0$, $\sqrt{(t+\cos\beta)^2 + (a\sin\beta)^2}$ (triangle OAB). Hence:

$$F_{Spring} = K\left(\sqrt{(t+\cos\beta)^2 + (a\sin\beta)^2}\right)(1-\cos\theta) = KU(\theta) \quad (12)$$

From (12) and (11):

$$K = \frac{1}{t\cos\theta(U(\theta))}\left((F'_N - mg)a\cos\left(\theta + \sin^{-1}(\frac{t}{a}\cos\theta)\right) - f_s H\right) = G(\theta) \quad (13)$$

To acquire pure rolling, we have:

$$K = max(G(\theta)) \quad (14)$$

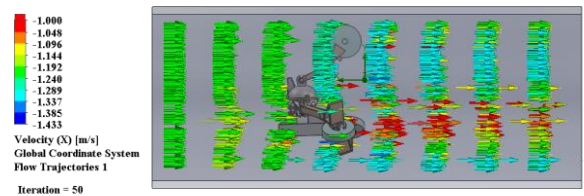

**Figure 5.** Flow Simulation Environment in SolidWorks.

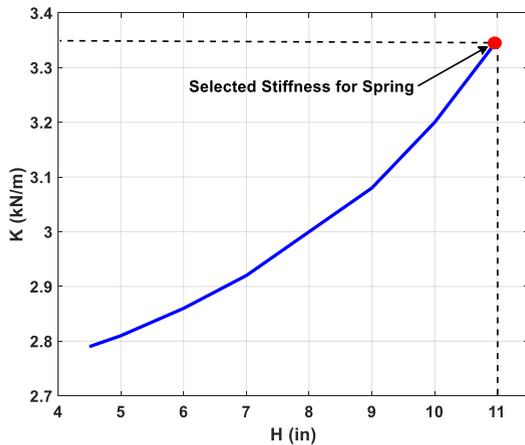

**Figure 6.** Stiffness Value Calculation for Pure Rolling of Wheels during Operation in Different Sizes of Pipe in Size-adaptability Range of the Robot and Selected Value for Stiffness.

In addition, (15) that is coulomb friction force relation needs to be satisfied:

$$f_{s(\max)} = \mu_s F'_N \quad (15)$$

where $\mu_s$ is friction coefficient and is 0.8 for pipe wall and wheel contact [9]. We need to find a value for $f_{s(\max)}$. It needs to be greater than the maximum traction force that each wheel needs to provide and to calculate the maximum traction force, we performed flow simulation with computational fluid dynamics (CFD) work in SolidWorks to compute the maximum resistive force against the robot. In this scenario, the robot moves with 50 cm/s velocity in opposite direction of flow motion with 70 cm/s velocity; the outer diameter of the robot is 22-in that is its maximum diameter for it and the line pressure is 100 kPa (standard line pressure). Table 1 shows the simulation specifications and Fig. 5 shows the simulation environment. The colored lines show the velocity counter around the robot in pipe. In this scenario, the drag force is computed around 26.1 N that the wheels should provide. Since the geometry of the robot is symmetric, each wheel needs to provide one third of the maximum drag force (i.e. 8.7 N). Hence the value for traction force is 8.7 N that is equal to $f_{s(\max)}$ and based on (15), $F'_N = 11$ N. We calculated the required stiffness in each pipe diameter based on (14) and the results are shown in Fig. 6. To cover all pipe diameters, we selected the maximum value for $K$ that is the required stiffness in 22-in diameter with value of 3.35 kN/m. Hence, we calculated the value for the spring stiffness based on the maximum drag force that is computed in the CFD work and provide sufficient normal force for the wheels for pure rolling. The spring stiffness for all springs are equal.

TABLE I
FLOW SIMULATION SPECIFICATIONS

| Parameter [unit] | Value |
|---|---|
| Fluid Type | Water |
| Robot Velocity [cm/s] | 50 |
| Flow Velocity [cm/s] | 70 |
| Pipe Diameter [in] | 22 |
| Relative Direction of Robot and Flow Motion | Opposite |
| Line Pressure [kPa] | 100 |

### 2) BATTERY:

The other important factor for autonomous robot is battery capacity. We calculated the battery capacity with (16):

$$C = \frac{3P.h}{V_n} \quad (16)$$

where $C$ (A.h), $P$ = 20W, $h$ (hour), and $V_n$= 12 V are battery capacity, gear-motor power, assumed operation duration, and gear-motor operating voltage. In this method, an operation procedure is assumed and battery capacity is calculated and a battery is selected. Then, the drawn current from the battery by gear-motors (at the operating point for traction force of 8.7N in the CFD work) is defined. The discharge time of the battery at this drawn current and voltage (12 V) is defined and compared with the assumed operation duration. If they are approximately equal, the battery capacity and the operation duration are realistic, otherwise, we need to consider another operation duration and repeat the process. In our system, the battery capacity is calculated around 15 A.h and the operation duration is around 3 hours.

### 3) DISCUSSION

We characterized the robot's components based on the extreme operation conditions to have sufficient normal forces between the pipe wall and the wheels for pure rolling of the wheels. In addition, we characterized the battery capacity based on the extreme operation condition and defined an operation duration for the robot. Hence, the robot does not run out of power during operation.

## IV. WIRELESS SENSOR MODULE

Our in-pipe robot is self-powered and needs communication with base station to exchange sensors' data and motion control commands through wireless communication. In this application, the robot moves inside pipeline and the transceiver(s) are located outside pipe (above ground) a few meters away of the robot (we will later explain about transceivers location in this paper). The robot switches its communication between transceivers during operation. Hence, there is a need for fast discovery between the robot and the transceivers. Also, we need a bi-directional wireless system that facilitates data transmission on both sides (i.e. from the robot to the base station and from the base station to the robot). However, wireless communication in underground applications is challenging since the environments of water, pipe, and soil attenuate radio signals (i.e. high path loss) [10], and also the communication channel is dynamic in which the volumetric water content and also the sand-clay composition in soil are variable and affect the path loss in soil [10]. To mitigate high path loss, low-frequency carrier signals for wireless communication is desired [10].

Considering the aforementioned requirements, we propose a wireless sensor module based on radio frequency identification (RFID) technology that can operate in low carrier frequencies [27]. To this aim, we use CC1200 from Texas Instruments (TI) Inc© as physical layer of the RFID module that facilitates bi-directional communication and works in five sub-1GHz frequency bands (i.e. 169, 434, 868, 915, and 920 MHz) [28]. Also, it has 128 byte data FIFO on transmitter (TX) and 128 byte data FIFO on receiver (RX) sides that facilitates high data throughput in our application [29]. The functions of the CC1200 (e.g. data packet creation, transmission of data, reception of data, etc.) are controlled with the host MCU [30]. Our wireless sensor module comprises of the RFID module that is connected to the five sensors for up to five parameters measurement in water that is improvement with regard our team's previous work [5] (one parameter measurement) and also comparable with [31] that perform five parameters measurement.

TABLE II
WIRELESS SENSOR MODULE SPECIFICATIONS

| Parameter [unit] | Value or Description |
| --- | --- |
| Physical Layer | CC1200 |
| Carrier Signal Frequency [MHz] | 434 |
| Antenna | Taoglas TI.10.0111 (SMA Mount) |
| MCU | ATMEGA2560 |
| Number of Sensors | Up to 5 |
| Unit 1 | Micro-pump System |
| Unit 2 | Motion Control |

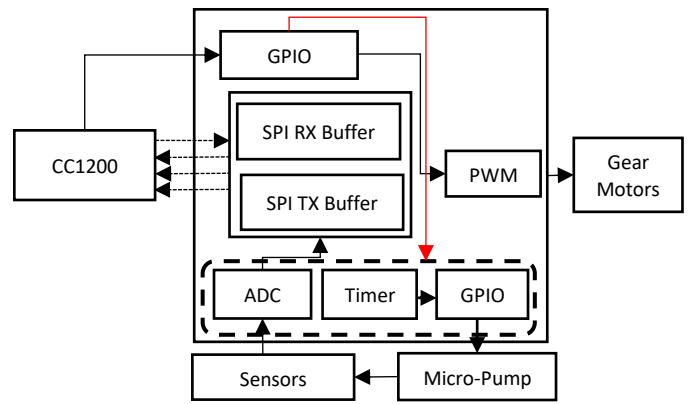

**Figure 7.** Block Diagram for the Wireless Sensor Module.

The physical layer (i.e. CC1200) triggers one interrupt on the MCU's pin (GPIO in Fig. 7) upon reception and transmission of a data packet. We use this feature of the physical layer to control the function(s) of two units: micro-pump system and the motion control unit. The micro-pump system provides water samples for the sensors that need water samples for measurement (mp6, Bartels Mikrotechnik). When an interrupt occurs, the micro-pump starts circulating water in the sensors and the ADC of the MCU starts taking sample (dashed rectangular in Fig. 7). Also, the interrupt is connected to the motion of control unit (Pulse Width Modulation (PWM) signals for the gear-motors). We will explain the details about the micro-pump and the motion control units, later in this paper. The wireless system in this system works in 434 MHz carrier frequency, the MCU is ATMEGA 2560, the antenna type is Taoglas TI.10.0111 with SMA mount, that are printed with other parts of the electronic components a circular printed circuit board (PCB) with 2.94-in diameter and located in the central processor (see Fig. 2c). Table II shows the specifications of the wireless sensor module.

*A. Functionality of the Wireless Sensor Module: Bi-directionality. Fast Discovery Capability, and Maximum Throughput of the Wireless Sensor Module*

We mentioned in earlier that we need a bi-directional wireless communication, fast discovery capability between the robot and the transceivers, and high data throughput. In this section, we evaluate the performance of the proposed wireless sensor module in terms of bi-directionality, connection pick capability, and also measure the maximum data throughput. To this aim, an experiment setup is designed that is shown in Fig. 8. In this setup, five FlexiForce sensors ([c] in Fig. 8) are connected to the PCB of the robot ([f] in Fig. 8) and powered with the power supply ([d] in Fig. 8). The PCB is powered with the battery. Two transceiver modules ([a] and [b] in Fig. 8) are designed that each one includes an MCU Launchpad and CC1200 evaluation module. In each experiment in the following, a specific

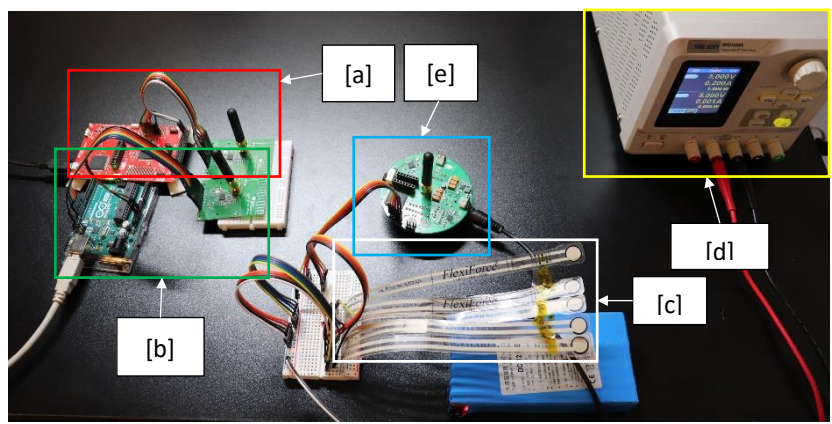

**Figure 8.** Experiment Station for the Wireless Sensor Module Functionality. [a] Transceiver Module 1. [b] Transceiver Module 2. [c] Five Force Sensors. [d] Power Supply to Power Sensors. [e] PCB.

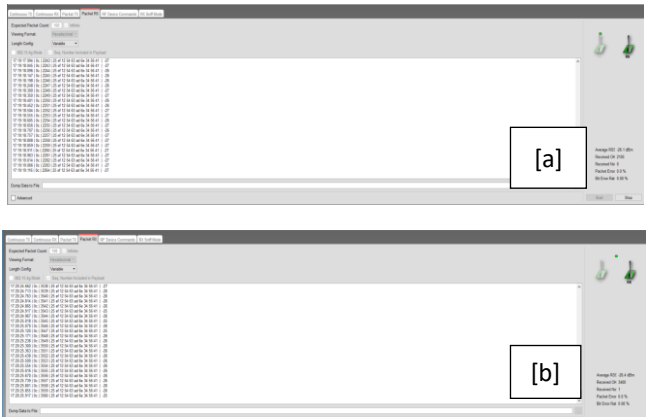

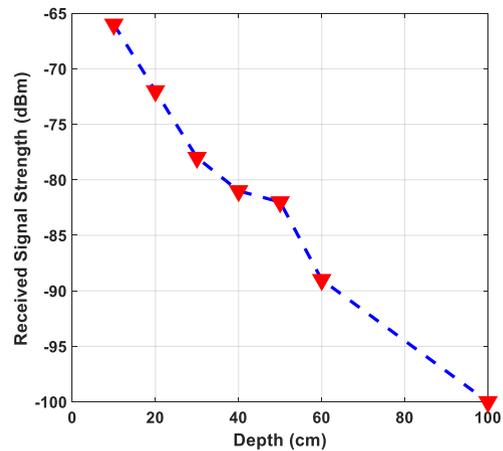

**Figure 9.** Graphical User Interface (GUI) for Monitoring the Received Data Packets for [a] Transceiver Module 1, [b] Transceiver Module 2.

**Figure 11.** Received Signal Strength (RSS) at different vertical distances of transceiver 1 and transceiver 2.

architecture of the setup in Fig. 8 is implemented to validate the desired functionality.

### 1) BI-DIRECTIONALITY

To evaluate the bi-directionality capability of the wireless sensor module to the transceiver, we performed two experiments: in the first experiment, the data packets are created in the wireless sensor modules and transmitted to the transceiver module 1. We sent 100 data packets and since an interrupt occurs on a designated pin upon reception of a data packet in transceiver module 1, it is supposed to see 100 interrupts on the pin in this experiment. We did and verified 100 interrupts on the 100 data packets on the transceiver module 1 with logic analyzer.

In another experiment, the data packets are sent from the transceiver module 1 to the PCB and the same procedure is repeated and the same number of interrupts on the wireless sensor module's MCU and the transmitted data packets are verified. Hence, the interrupt capability of the wireless sensors module facilitates knowledge about the transmission and reception of data packets and also the number of data packets. Hence, we have a bi-directional wireless communication system in our application.

### 2) CONNECTION PICK CAPABILITY

The robot moves in pipe and the transceivers are located outside pipe. We have multiple transceivers in our application (we will elaborate more about this in this paper) and it is important for the wireless sensor module on the robot to easily establish communication with the transceivers when it reaches to the read range of the transceivers.

To evaluate the capability of fast discovery between the wireless sensor module and the transceivers, the wireless sensor module sends data packets to Transceiver Module 1, so we have communication between them and a graphical user interface (GUI) monitors the received packets (see Fig. 9a). During this transmission, transceiver module 2 at receive mode is added to the network and it is expected that the second transceiver also receives the same data packets as well. We realized that the transceiver 2 receives the same data packets as the transceiver 1 (see Fig. 9b). Hence, there is fast discovery between the transceivers and the wireless sensor module in this application.

### 3) MAXIMUM THROUGHPUT

The maximum data rate in our wireless system was measured to be around 120 kbps that is greater than similar wireless underground communication system [32]–[34] and also sufficient for wireless underground applications [10].

### 4) SIGNAL PENETRATION IN WATER ENVIRONMENT

To evaluate the performance of the developed sensor module, we designed a prototype of the wireless system with CC1200 evaluation module in 434 MHz, Arduino MEGA 2560 REV3 (MCU), and five force sensors that are connected to the ADC channels of the MCU. The dynamic range of the force sensors are the same as the water sensors. The prototype is powered with battery, is located in a water resistant bag, and submerged in water (see Fig. 10). The wireless sensor module creates the data packets from the sensor measurements and sends them wirelessly with 14 dBm that is the maximum transmit power that the physical layer can transmit data [28] and the customized antenna (Taoglas TI.10.0111). In the first experiment, horizontal distance between the transceiver 1 and transceiver 2 is zero ($H = 0$) and vertical distance is increased ($D \uparrow$). The results in Fig. 11 shows that the RSS is decreased from -66 dBm at 10 cm to -82 dBm at 60 cm and comparing to the -80 dBm that is the threshold power for correct realization of radio signals based on [33]. In another experiment, the transceiver 1 is submerged in different values of depths and horizontal distances of the transceiver (see Fig. 12). The results show that RSS does not change when horizontal distance between the transceivers increased, however, it changes dramatically when the depth of transceiver 1 changes. Hence the distance of the transceivers in the air does not affect the RSS, however, the distance in the water medium affects the RSS substantially and we can conclude that the read range of the developed wireless module with the customized antenna is around 40 cm (15.75 in) that is aligned with the size adaptability of the robot for the pipe radius (4.5 in-11 in); in real application, the robot moves in pipe with water. The antenna of the wireless sensor module is located in the central processor that is located in the center of pipe full of water (see Fig. 13).

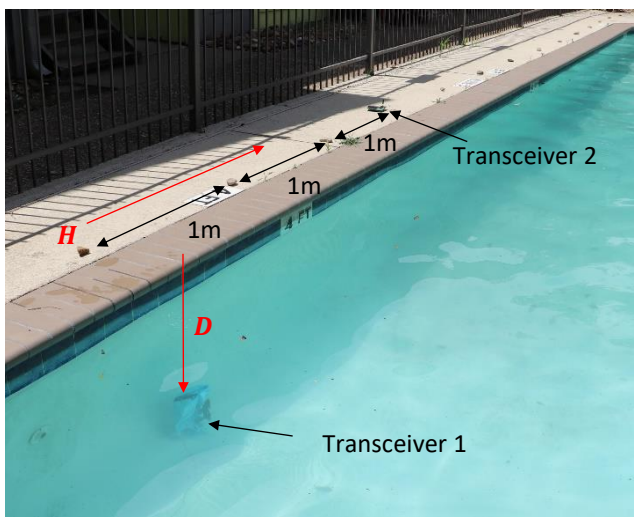

**Figure 10.** The strength of the radio signal at water medium. Transceiver 1 is located in a water resistant bag and submerged in a bucket of water. The depth of water is 1 m and the transceiver 2 is located on the ground a moves horizontally. $D$ is the vertical distance of transceivers and $H$ is the horizontal distance between receivers.

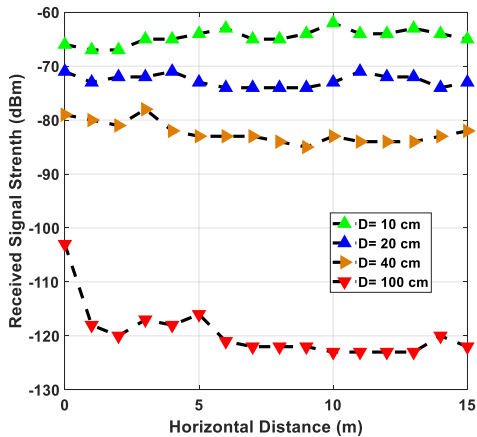

**Figure 12.** Received Signal Strength (RSS) at Different Horizontal and Vertical Distances of Transceiver 1 and Transceiver 2.

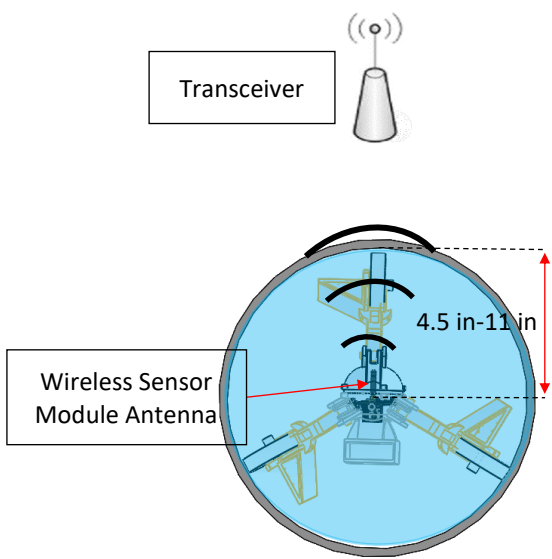

**Figure 13.** The Location of the Antenna in the Central Processor and the Distance from the Transceiver.

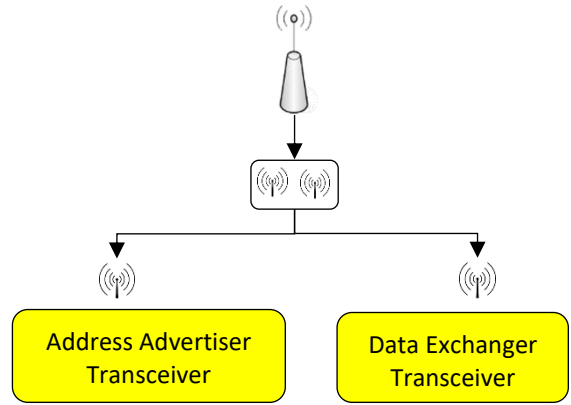

**Figure 14**: [a] Each relay node (RN) comprises two parts: Address Advertiser Transceiver (AAT) and data exchange transceiver (DET). [b] A Setup for One Relay Node.

In this section, we developed a bi-directional wireless sensor module that works based on RFID module and customized antenna. It facilitates multi-parameter (five parameters) measurements and transmissions with maximum data rate of 120 kbps. The read range of the wireless sensor module is around 40 cm in water medium that is aligned with the size-adaptability of the robot.

## V. OPERATION PROCEDURE

So far, we have a size-adaptable in-pipe that is equipped with a bi-directional wireless communication system for pipe inspection in harsh environments in which signal attenuation is high. In this section, we propose an operation algorithm that synchronizes the wireless communication system with the motion control algorithm and data acquisition unit of the robot.

Our objective is to facilitate smart navigation and wireless data transmission for the robot during operation. First, we explain the system hardware and then the operation procedure.

### A. System Hardware

Our proposed wireless robotic network includes:

- One wireless sensor module that is mounted on the robot and moves with it during operation.
- Some fixed relay nodes (RN) above ground.

The wireless sensor module on the robot synchronizes the wireless sensor, micro-pump system, and the motion control units and we explained about it earlier in this paper. We call the transceiver on the robot, moving transceiver (MT) for brevity. The RNs include two radio transceivers that are connected to the host MCU. One transceiver always advertises an address signal on the air and the other one exchanges data with the robot. We call the transceiver that advertises the signal, address advertiser transceiver (AAT), and the transceiver that exchange data, data exchange transceiver (DET) (see Fig. 14). The RNs are located at special configurations of the pipeline like bend, wyes, and T-junction. Fig. 15 shows the overall view of the proposed wireless robotic sensor that the robot moves inside pipeline under ground and the RNs are located at special configurations of the pipeline.

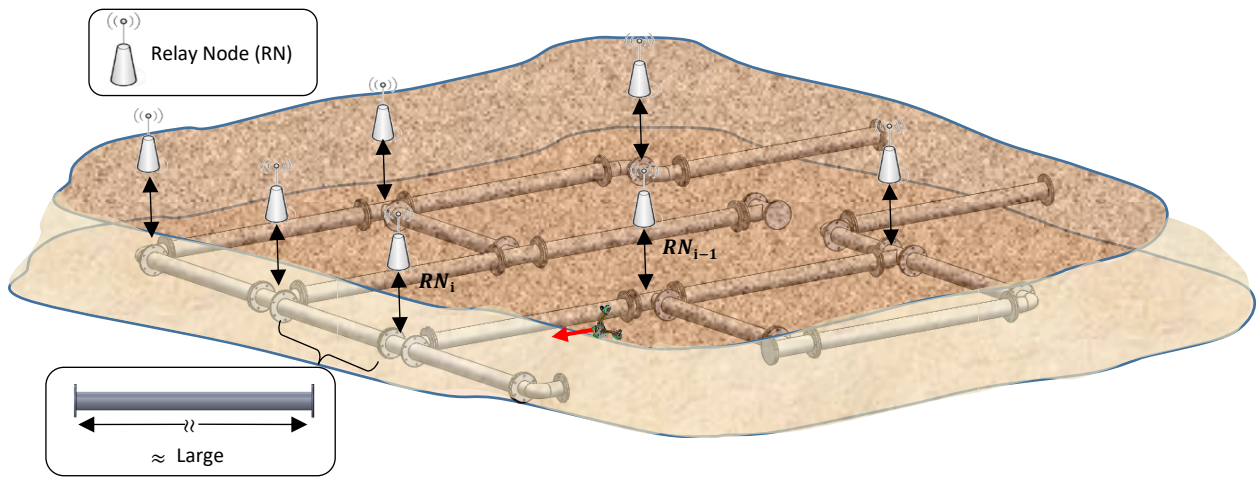

**Figure 15.** Overall View of Our Proposed Robotic Network. The robot moves inside pipe underground and the RNs are located at special configurations of the pipelines like bends and T-junctions. The distance between the RNs is large.

### B. Operation Procedure

The operation procedure of the robotic divides into five parts:
1) Motion in Straight Path with No Wireless Communication: The robot moves in a straight path between two consecutive RNs ($RN_{i-1}$ to $RN_i$) and it does not have wireless communication (see Fig. 16).

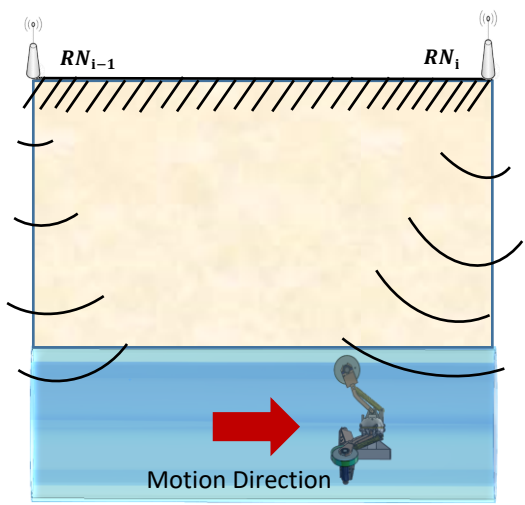

**Figure 16.** Part 1 of the Proposed Operation Procedure: The robot moves from $RN_{i-1}$ to $RN_i$ and the robot does not have wireless communication with relay nodes.

2) Establishing Communication between the Robot and $RN_i$:
The robot establishes wireless communication with $RN_i$ when the robot arrives in the radio range of $RN_i$.
3) Sensor Measurement, Processing, and Wireless Transmission from the Robot to the $RN_i$:
The robot starts sensor measurements, processes and transmits them to the $RN_i$.
4) Wireless Transmission from the $RN_i$ to the Robot:
The $RN_i$ transmits the motion control command to the robot.
5) Change of Direction of the Robot:
After the robot received the motion command signal, it changes its direction and starts moving in a straight path until it arrives at the radio range of the $RN_{i+1}$.

The operation procedure of the robot is in a way that the robot switches between different control algorithm modes and data transmission directions. In the following, we explain each part of the procedure in detail.

#### 1) MOTION IN STRAIGHT PATH WITH NO WIRELESS COMMUNICATION:

In this phase of the operation, the robot moves in a straight path and does not have wireless communication with RNs. The motion controller in this phase stabilizes the robot and makes it track a constant velocity. To have a measure of stabilization for the robot, we define rotation around the y-axis ($\phi$), z-axis ($\psi$), and their derivatives, $\dot{\phi}$ and $\dot{\psi}$ (see Fig. 17). The controller effort for the stabilization task is to keep the stabilizing states of the robot, $[\phi \quad \psi \quad \dot{\phi} \quad \dot{\psi}]$ at zero [9], [25]. To this aim, we designed a controller that is the linear quadratic regulator (LQR) and stabilizes the robot. Also, the robot needs to track a desired velocity during operation. To this aim, another controller is proposed which is based on proportional-derivative-integral (PID) controller. If all the three wheels of the robot have equal angular velocity, the linear velocity of the robot is calculated based on one wheel's angular velocity. So, to have a defined linear velocity, we proposed three PID controllers to control the velocity of each wheel. The LQR and PID controllers are combined to fulfill both requirements in a straight path. Fig. 18 shows the controller algorithm diagram in this phase [9], [25]. The observer in the controller algorithm fuses the data from an inertial measurement unit (IMU) which is placed in the central processor and three encoders at the end of the wheels. The encoders measure the angular velocity of the wheels. Hence with this architecture, we have a sense of orientation of the robot and odometry for it. We validated the performance of the controller in this phase with experimental results. To this aim we performed four iterations with different desired linear velocities, $V_d$, and initial values for $\phi$, $\phi_0$ and $\psi$, $\psi_0$. In Table III, the specifications of each iteration are defined. The performance of the controller in this phase is shown in Fig. 19. For the linear velocity, the robot starts from zero velocity and reaches to the desired velocity. Also for the stabilizing states, the initial deviations are cancelled and kept at that point with small fluctuation (see Fig. 19).

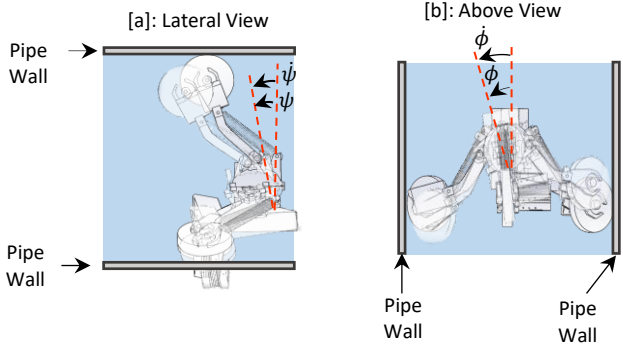

**Figure 17.** Stabilizing States of the Robot.

TABLE III
EXPERIMENT SPECIFICATIONS IN FOUR ITERATIONS FOR PHASE 1 OF THE MOTION CONTROLLER.

| Iteration | $V_d$ (m/s) | $\phi_0$ (degree) | $\psi_0$ (degree) |
|---|---|---|---|
| 1 | 0.1 | +14 | -15 |
| 2 | 0.2 | -13 | -11 |
| 3 | 0.3 | -9 | +5 |
| 4 | 0.35 | -3 | +3 |

The results show the LQR-PID controller is able to stabilize the robot in pipeline and makes the robot track a desired velocity [9].

2) ESTABLISHING COMMUNICATION BETWEEN THE ROBOT AND $RN_i$:

In pipeline underground, the robot is moving in a straight path inside pipe from $RN_{i-1}$ toward $RN_i$. On the ground in $RN_i$, one transceiver on the $RN_i$ is in transmit mode and advertises the $RN_i$'s location order continuously. The location order is the address of the $RN_i$ in the network that we call it "relay node address" or RNA in short. For example, the $RN_1$ is the first relay node in the network in which the robot approaches to and the RNA for it (i.e. $RN_1$) is 1. The $RN_2$ is the second one in the network and its RNA (i.e. $RN_2$) is 2, and so on. Since each RN advertises a unique RNA, the possible overlap between radio signals of multiple RNs does not confuse the robot (Fig. 15).

On the robot side, the robot needs to distinguish the right RN in the network. This way it can switches to the right control algorithm that facilitates efficient motion during operation. In other words, the robot needs a map of the operation path. Our solution to give the robot a sense of the map of the pipeline is to put the RNs in the special configurations of the pipeline network and give the robot an array that contains the RNAs and their orders in the network:

$$\mathbf{RNA} = [RNA_1 \quad RNA_2 \quad ... \quad RNA_n] \quad (17)$$

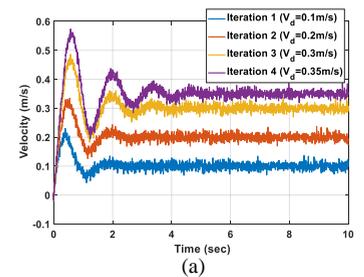

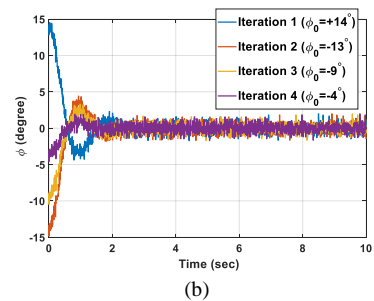

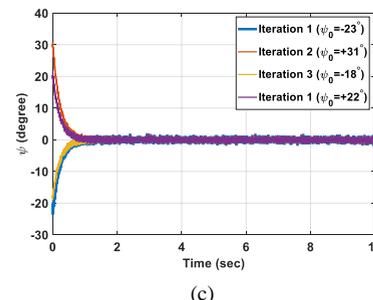

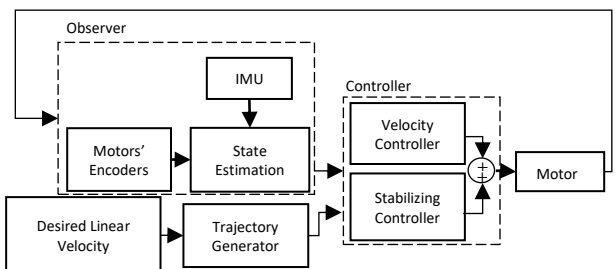

**Figure 18.** The controller to stabilize the robot and track the desired velocity in straight paths.

**Figure 19.** The proposed LQR-PID based controller performance in straight path based on experimental results. Four iterations are done to validate the performance of the controller. (a) Linear velocity of the robot along pipe axis. (b) $\phi$ (degree). (c) $\psi$ (degrees).

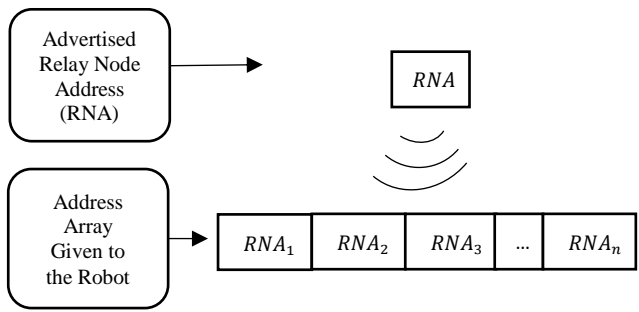

**Figure 20.** Advertised RNA and the address array that is programmed in the robot firmware.

In (17), the $RN_i$ reveals the information about the configuration type. The configuration type can be 90° bend, 45° bend, 135° bend, T-junction, etc. The robot is programmed in a way that knows the configuration shape associated with specific RNA. For example, the configuration associated with $RN2$ is a 90°-bend. Hence, the control algorithm chooses the control phase that steers the robot in a bend shape pipe. Fig. 20 shows the advertised RNA and the array in the robot's firmware. Algorithm 1 defines configuration type of the pipe where the robot is located. The motion controller [9]. The motion controller in this part of the procedure is stabilizer controller in which the performance of the controller is similar to Fig. 19 with the difference that the linear velocity is zero in this phase. Hence, we localize the robot in the network with the wireless communication and the map of the operation in the robot's firmware.

3) SENSOR MEASUREMENT, PROCESSING, AND WIRELESS TRANSMISSION FROM THE ROBOT TO THE $RNI$:

So far, the robot has received a RNA, stopped moving, and defined the configuration type. At this point, the micro-pump system activates and provides water samples for the onboard water quality monitoring sensors. The quality monitoring sensors are chemical sensors (the number of the sensors depends on the application.) that are designed and developed by our team [5]. These chemical sensors operate based on chemical reactions and their output converges to a specific value after a while. We consider one minute for settling time

**Algorithm 1: Configuration Type Determination**

1. Interrupt on Wireless Pin?
2. done =
3. $i=1$
4. while (!done):
   If ($RNA_i == RNA$)
     done='True'
   end
   $i = i + 1$.
   end
5. Configuration type is defied.

**Algorithm 2: Phase 3 of the Operation.**

1. Micro-pump Activation.
2. Sensors' Output Settlement (One Minute).
3. Start Sensor Measurement.
4. Sensor Measurement Processing.
5. Data packet Creation with Processed Sensors' Data.
6. Update of the MT's Payload with Data Packet.
7. Data Packets |))
8. Done Transmission (DT) Signal |))
9. TX Mode → RX Mode.

based on [5]. The sensors are connected to the host MCU. After, their output is settled, the analog to digital conversion (ADC) unit of the MCU starts sensor measurements. The MCU then processes raw sensor data to compute the real value of the target analyte. The processing includes filtering noisy data, converting current to voltage, and applying the associated chemical formula to calculate the real value for the target analyte. At this moment, the processed sensor measurements are ready. Before sending the data to the $RN_i$, we need to create a data packet that complies with the communication protocol. With a packet generator function, a data packet is generated that is an array which includes the preamble bytes, packet length, the packet counter, and processed sensor data. The payload of the MT is then updated with the data packet, and then the MCU transmits packet to the $RN_i$. The transceiver on the $RN_i$ which is in receive mode, receives the packet. After the MCU transmits all data packets, it sends a Done Transmission (DT) signal to the $RN_i$, and the MT goes to receive mode. Algorithm 3 shows the procedure to start measurement and create data packets.

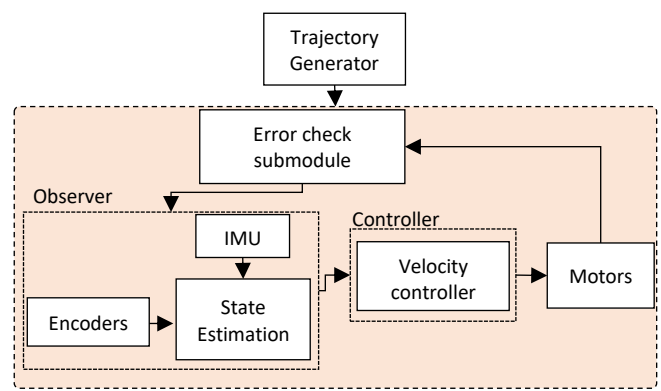

**Figure 21.** The Motion Control Algorithm for Non-straight Paths.

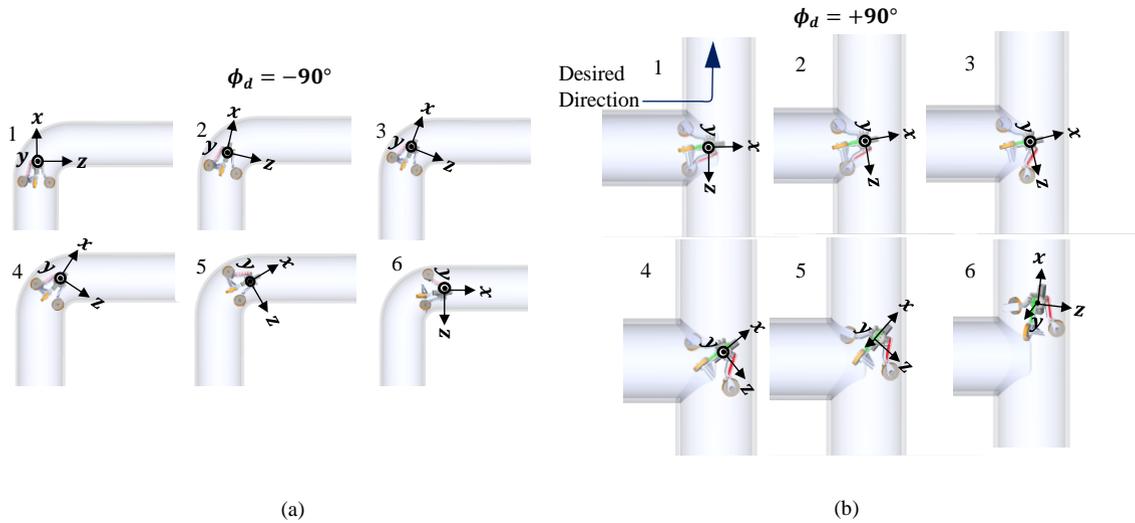

Figure 22: DAMS-MATALB Co-simulation Results for (a) 90° Bend and (b) T-junction based on Non-straight Controller

4) WIRELESS TRANSMISSION FROM $RN_i$ TO THE ROBOT:

In this phase, the $RN_i$ sends a motion control command to the MT. Based on the configuration type that is already defined in the part 2 of the operation, the robot chooses the control phase designed for non-straight configuration that steers the robot to the desired direction. The controller includes a trajectory generator block, an observer block, an error-check sub-module, and a velocity controller based on three PID controller. The trajectory generator creates differential motion in which different angular velocities for the wheels, change the direction of motion of the robot. The error-check submodule monitors the amount of rotation around the desired axis and allows the controller in the non-straight phase to continue until the defined amount of rotation is acquired. Fig. 21 shows the control algorithm for the non-straight configuration. To evaluate the performance of this phase of the controller, we modeled the robot in AMDAS software which is a multibody dynamic simulator and implemented the controller in MATLAB Simulink environment. Then we linked the control plant in MATLAB and the simulated system in ADAMS software, and co-simulated the robot-controller [26]. Fig. 22 shows the motion sequences of the robot in bend and T-junction. The results in Fig. 22 prove that the motion control phases for the robot in T-junctions and bends can facilitate smooth motion for the robot in bends and T junctions. We repeated our simulations and validated the robot can cover bend with diameters range 9-in to 22-in and also T-junctions with diameters range from 9-in to 15-in. Once the rotation is completed, the robot is located in a straight path and continues moving in a straight path. The control phase here is again the stabilizer-velocity tracking controller in Fig. 18.

5) CHANGE OF DIRECTION OF THE ROBOT:

The robot switches to the non-straight controller phase and changes its direction to the desired direction. The robot communication with $RN_i$ stops during rotation and $RN_i$ is removed from the service.

Fig. 23 shows switching between different motion control phases and communication direction modes during operation.

We synchronized the wireless communication setup and the motion control algorithm in this work. Fig. 24 shows the overall view of the synchronized wireless control for the operation procedure of the robot. Hence, the robot navigates between different configurations of the pipelines with the synchronization of the wireless sensor module and the multi-phase motion control algorithm.

C. Experiment

In this part, we evaluate the performance of the developed operation procedure with experiment. To this aim a relay node is proposed with two CC1200 evaluation modules in 434 MHz that are connected to the two MCU Launchpads

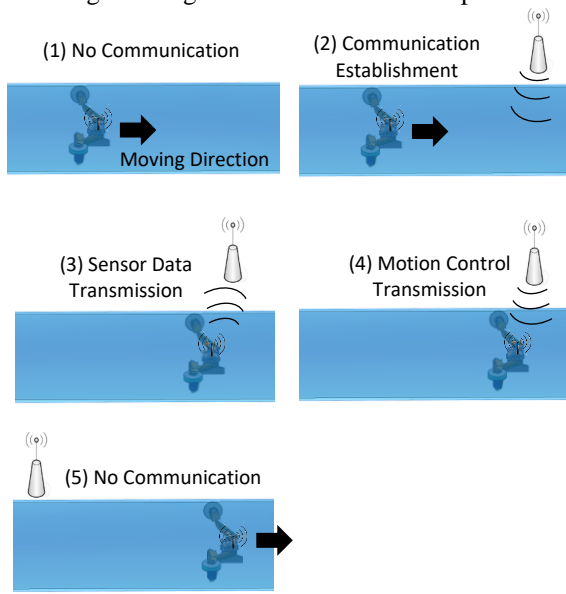

Figure 23. Wireless Communication Switching in the Proposed Operation Procedure for the Wireless Robotic Network.

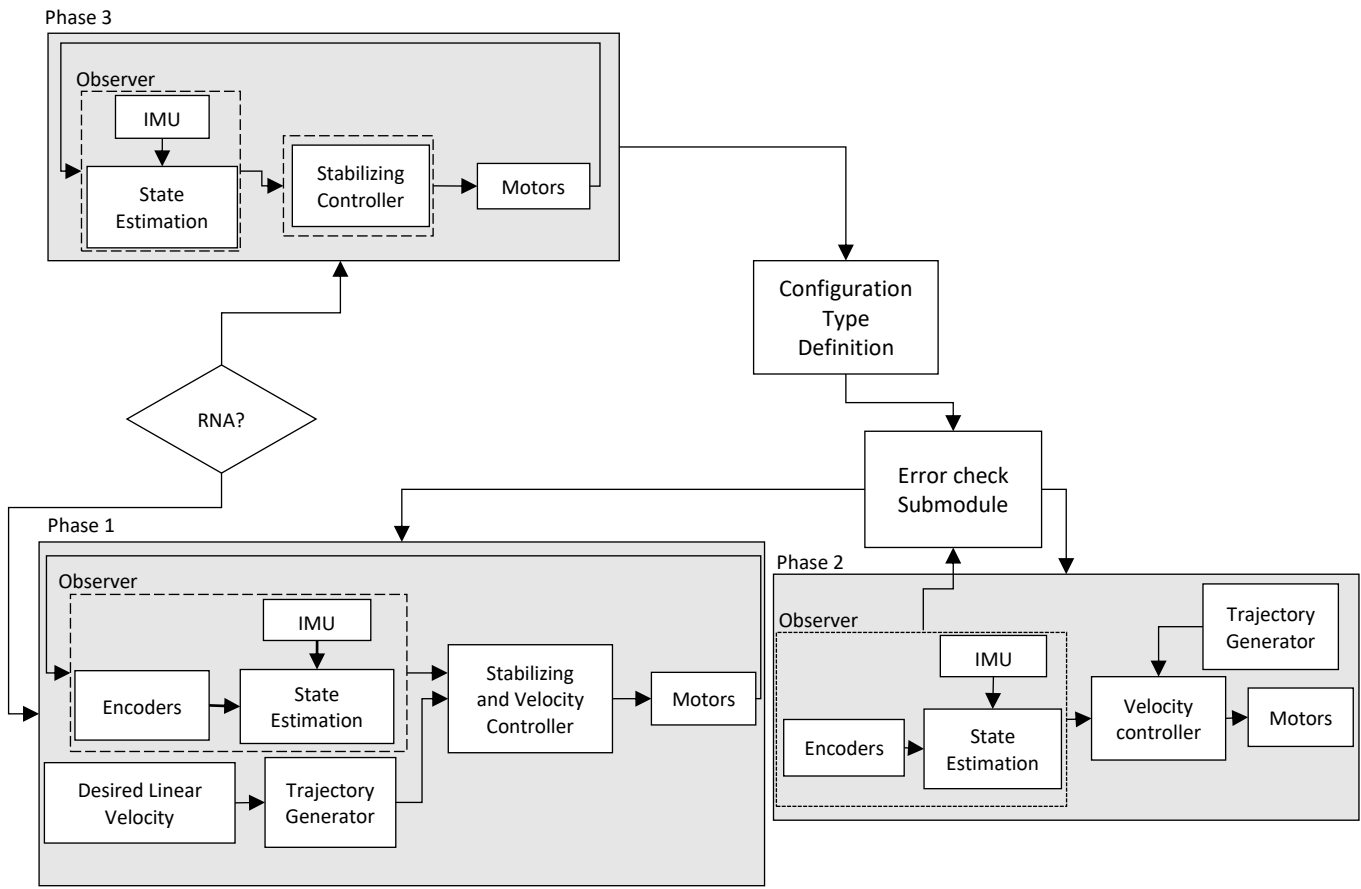

**Figure 24.** Synchronized Wireless Communication System with the Multi-phase Motion Control Algorithm. Switching between different phases of the motion control algorithm is performed with wireless system.

(see Fig. 25). The robot is located in a 14-in diameter pipe. Our goal is to investigate the functionality of the wireless control and navigation method. To this aim, the relay node sends starts command to the robot and the robot starts motion with the phase 1 of the controller (stabilizer-velocity tracking controller) and move inside the pipe with 10 cm/s. The robot has the map of the operation as **RNA** = [1] as we have just one junction in our experiment. The relay node sends the $RNA = 1$ when the robot reaches to the end of the pipe and it is expected that the robot stops at this location once it received the $RNA$. We performed the experiment and the results are shown in Fig. 26 in which the sequences of motion aare shown from 1 to 4. The robot first stabilizes itself at the beginning of the motion and reach to the desired velocity (i.e. 10 cm/s) and moves with stabilized motion in the pipe. At the end of the motion, the relay node sent the $RNA = 1$ and upon transmission, the robot stopped there with the stabilized configuration. Hence, the robot switched from phase 1 to phase 2 of the motion controller and the developed wireless control is verified.

1) LIMITATION

In our experiments, we found out that the time at which the AAT sends the $RNA$ is extremely important as the robot stops with the stabilized configuration once it receives the $RNA$. In other words, switches to another phase of motion control algorithm. In some experiments, the robot stopped near the end of the pipe (desired location) while in others experiments, the robot stopped at other undesired locations. This is because the radio signal propagates spatially in air and the robot may receive it at any location. In our experiment, this phenomenon is higher as water medium that limits the read range of the radio signal is not present and in real applications this effect is less than in our experiment but still exist. We can address the issue by locating a rangefinder sensor in front of the robot and find the distance from the front obstacle and reach close enough to the junction with the stabilized motion (phase 1) in straight path before switch to phase 2 of the motion controller.

## VI. DISCUSSION AND CONCLUSION

The developed operation procedure facilitates localization and navigation in complicated configurations of pipelines for the in-pipe robots that was a challenge in this field. Also, the proposed method facilitates long-distance inspection for the robot; since the distance between the rely nodes are large. The resilient of the robot that is coined and discussed in [26] is considered in this robot in which the spring mechanism that provides friction force between the pipe wall and the wheels is characterized based on the extreme operating condition in in-service networks as well as the battery capacity of the robot. The spring mechanism prevents the robot from collapse during operation and the characterized battery ensures enough operation duration that is at least 3 hours in the conditions the robot moves with 50 cm/s against the water flow with 70 cm/s [9]. Even in case the robot fails during operation, it is easy to pinpoint the location of the failed robot with this method. We also validated the operation procedure in each step separately by experiment and simulations; the performance of the motion controllers are validated with experiments in straight path and simulation in non-straight configurations, the reliability of wireless communication system is validated to ensure it is able to penetrate harsh environments of water and pipe, bi-directionality, and fast discovery between the robot and the relay nodes. The sensors that measure the parameters of water have rather long response time (around one minute [5]) and we considered their response time in the procedure. The developed procedure enables data transmission during operation which is useful in quality monitoring in long inspections. We also experimented the wireless control to see its functionality to enable the robot for reliable motion in complicated configurations.

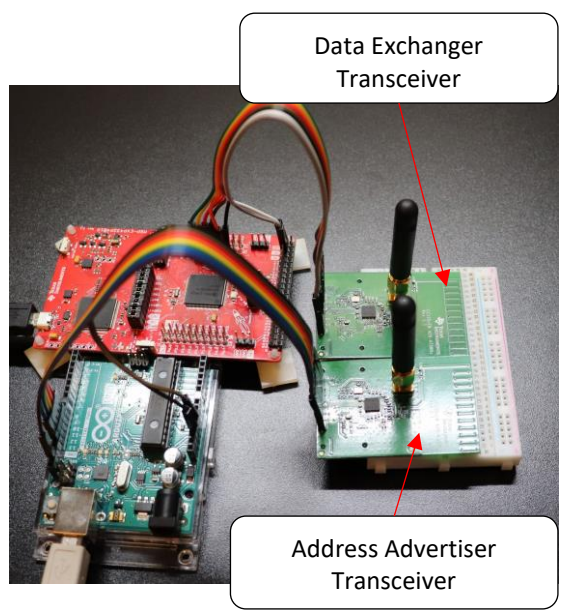

**Figure 25.** A Setup for One Relay Node. Each relay node (RN) comprises two parts: Address Advertiser Transceiver (AAT) and data exchange transceiver (DET). Two CC1200 evaluation modules are connected to two MCU Launchpads.

In our future work, we plan to do more field tests to evaluate the functionality of the robot and the developed controller in water networks.

### REFERENCES

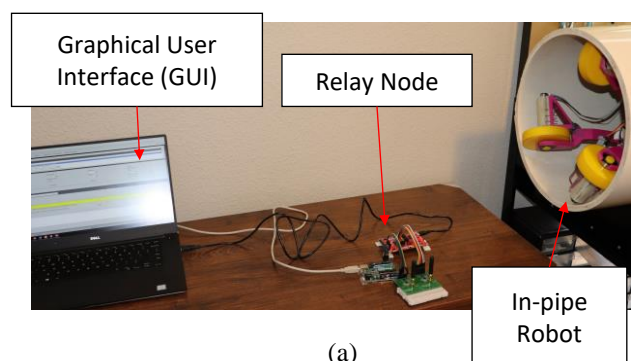

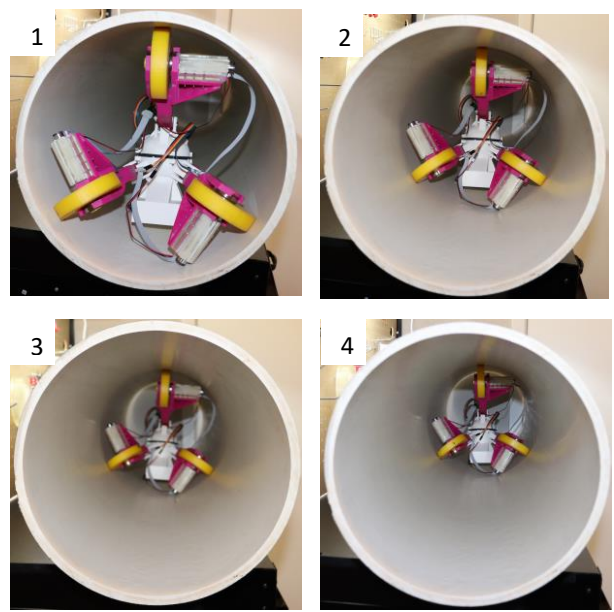

**Figure 26**. (a) Experiment Setup: The in-pipe robot is located in a 14-in diameter PVC pipe. The relay node is connected to PC and the sends the $RNA = 1$ once the robot received at the end of the pipe. (b) Sequence of motion of the robot. The robot starts moving in (1) with phase 1 of the motion controller (stabilizer-velocity tracking controller) and reach to 10 cm/s velocity. Once the robot arrived at the end of the pipe, the relay node sends $RNA = 1$ and the robot stops with stabilized configuration.

[1] A. L. Vickers, "The future of water conservation: Challenges ahead," J. Contemp. Water Res. Educ., vol. 114, no. 1, p. 8, 1999.
[2] E. Canada, "Threats to water availability in Canada." National Water Research Institute Burlington, 2004.
[3] D. Chatzigeorgiou, K. Youcef-Toumi, and R. Ben-Mansour, "Design of a novel in-pipe reliable leak detector," IEEE/ASME Trans. mechatronics, vol. 20, no. 2, pp. 824–833, 2014.
[4] R. Fletcher and M. Chandrasekaran, "SmartBall: a new approach in pipeline leak detection," in International Pipeline Conference, 2008, vol. 48586, pp. 117–133.
[5] R. Wu et al., "Self-powered mobile sensor for in-pipe potable water quality monitoring," in Proceedings of the 17th International Conference on Miniaturized Systems for Chemistry and Life Sciences, 2013, pp. 14–16.
[6] D. M. Chatzigeorgiou, K. Youcef-Toumi, A. E. Khalifa, and R. Ben-Mansour, "Analysis and design of an in-pipe system for water leak detection," in ASME 2011 International Design Engineering Technical Conferences and Computers and Information in Engineering Conference, 2011, pp. 1007–1016.
[7] K. Miyasaka, G. Kawano, and H. Tsukagoshi, "Long-mover: Flexible Tube In-pipe Inspection Robot for Long Distance and Complex Piping," in 2018 IEEE/ASME International Conference on Advanced Intelligent Mechatronics (AIM), 2018, pp. 1075–1080.

[8] Y. Qu, P. Durdevic, and Z. Yang, "Smart-Spider: Autonomous Self-driven In-line Robot for Versatile Pipeline Inspection," Ifac-papersonline, vol. 51, no. 8, pp. 251–256, 2018.
[9] S. Kazeminasab, A. Akbari, R. Jafari, and M. K. Banks, "Design, Characterization, and Control of a Size Adaptable In-pipe Robot for Water Distribution Systems," 2021.
[10] I. F. Akyildiz and E. P. Stuntebeck, "Wireless underground sensor networks: Research challenges," Ad Hoc Networks, vol. 4, no. 6, pp. 669–686, 2006.
[11] S. Kazeminasab, M. Aghashahi, Rouxi Wu, and M. K. Banks, "Localization Techniques for In-Pipe Robots in Water Distribution Systems," in 2020 8th International Conference on Control, Mechatronics and Automation (ICCMA), 2020, pp. 6-11, doi: 10.1109/ICCMA51325.2020.9301560.
[12] D. Wu, D. Chatzigeorgiou, K. Youcef-Toumi, S. Mekid, and R. Ben-Mansour, "Channel-aware relay node placement in wireless sensor networks for pipeline inspection," IEEE Trans. Wirel. Commun., vol. 13, no. 7, pp. 3510–3523, 2014.
[13] B. H. Lee and R. A. Deininger, "Optimal locations of monitoring stations in water distribution system," J. Environ. Eng., vol. 118, no. 1, pp. 4–16, 1992.
[14] D. Wu, D. Chatzigeorgiou, K. Youcef-Toumi, and R. Ben-Mansour, "Node localization in robotic sensor networks for pipeline inspection," IEEE Trans. Ind. Informatics, vol. 12, no. 2, pp. 809–819, 2015.
[15] S. Kazeminasab, M. Aghashahi, and M. K. Banks, "Design and Prototype of a Sensor-Based and Battery-Powered Inline Robot for Water Quality Monitoring in Water Distribution Systems," 2020.
[16] S. Kim and S. Kwon, "Robust transition control of underactuated two-wheeled self-balancing vehicle with semi-online dynamic trajectory planning," Mechatronics, vol. 68, p. 102366, 2020.
[17] N. Esmaeili, A. Alfi, and H. Khosravi, "Balancing and trajectory tracking of two-wheeled mobile robot using backstepping sliding mode control: design and experiments," J. Intell. Robot. Syst., vol. 87, no. 3, pp. 601–613, 2017.
[18] J.-X. Xu, Z.-Q. Guo, and T. H. Lee, "Design and implementation of integral sliding-mode control on an underactuated two-wheeled mobile robot," IEEE Trans. Ind. Electron., vol. 61, no. 7, pp. 3671–3681, 2013.
[19] Z. Wang et al., "Development of a Two-Wheel Steering Unmanned Bicycle: Simulation and Experimental Study," in 2020 IEEE/ASME International Conference on Advanced Intelligent Mechatronics (AIM), 2020, pp. 119–124.
[20] C.-C. Tsai, H.-C. Huang, and S.-C. Lin, "Adaptive neural network control of a self-balancing two-wheeled scooter," IEEE Trans. Ind. Electron., vol. 57, no. 4, pp. 1420–1428, 2010.
[21] J. Huang, Z.-H. Guan, T. Matsuno, T. Fukuda, and K. Sekiyama, "Sliding-mode velocity control of mobile-wheeled inverted-pendulum systems," IEEE Trans. Robot., vol. 26, no. 4, pp. 750–758, 2010.
[22] L. B. Prasad, B. Tyagi, and H. O. Gupta, "Optimal control of nonlinear inverted pendulum system using PID controller and LQR: performance analysis without and with disturbance input," Int. J. Autom. Comput., vol. 11, no. 6, pp. 661–670, 2014.
[23] S. Kim and S. Kwon, "Nonlinear optimal control design for underactuated two-wheeled inverted pendulum mobile platform," IEEE/ASME Trans. Mechatronics, vol. 22, no. 6, pp. 2803–2808, 2017.
[24] F. Raza, D. Owaki, and M. Hayashibe, "Modeling and Control of a Hybrid Wheeled Legged Robot: Disturbance Analysis," in 2020 IEEE/ASME International Conference on Advanced Intelligent Mechatronics (AIM), 2020, pp. 466–473.
[25] S. Kazeminasab, R. Jafari, and M. K. Banks, "An LQR-assisted Control Algorithm for an Under-actuated In-pipe Robot in Water Distribution Systems," 2021.
[26] W. J. Zhang and C. A. Van Luttervelt, "Toward a resilient manufacturing system," CIRP Ann., vol. 60, no. 1, pp. 469–472, 2011.
[27] K. Sattlegger and U. Denk, "Navigating your way through the RFID jungle," White Pap. Texas Instruments, 2014.
[28] "CC1200 CC1200 Low-Power, High-Performance RF Transceiver 1 Device Overview," 2013. Accessed: Sep. 29, 2020. [Online]. Available: www.ti.com.
[29] Z. Chu, F. Zhou, Z. Zhu, R. Q. Hu, and P. Xiao, "Wireless powered sensor networks for Internet of Things: Maximum throughput and optimal power allocation," IEEE Internet Things J., vol. 5, no. 1, pp. 310–321, 2017.
[30] S. Kazeminasab and M. K. Banks, "SmartCrawler: An In-pipe Robotic System with Novel Low-frequency Wireless Communication Setup in Water Distribution Systems." TechRxiv, Jan. 2021, doi: 10.36227/techrxiv.13554197.v1.
[31] A. U. Alam, D. Clyne, H. Jin, N.-X. Hu, and M. J. Deen, "Fully Integrated, Simple, and Low-Cost Electrochemical Sensor Array for in Situ Water Quality Monitoring," ACS sensors, vol. 5, no. 2, pp. 412–422, 2020.
[32] H. Guo, Z. Sun, and C. Zhou, "Practical design and implementation of metamaterial-enhanced magnetic induction communication," IEEE Access, vol. 5, pp. 17213–17229, 2017.
[33] Z. Sun, P. Wang, M. C. Vuran, M. A. Al-Rodhaan, A. M. Al-Dhelaan, and I. F. Akyildiz, "MISE-PIPE: Magnetic induction-based wireless sensor networks for underground pipeline monitoring," Ad Hoc Networks, vol. 9, no. 3, pp. 218–227, May 2011, doi: 10.1016/j.adhoc.2010.10.006.
[34] S. Kisseleff, I. F. Akyildiz, and W. H. Gerstacker, "Throughput of the magnetic induction based wireless underground sensor networks: Key optimization techniques," IEEE Trans. Commun., vol. 62, no. 12, pp. 4426–4439, 2014.

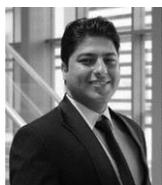

**Saber Kazeminasab** (S'18) is a PhD student in the Department of Electrical and Computer Engineering, Texas A&M University, College Station, TX, USA. He received the B.Sc. degree from the Iran University of Science and Technology, Tehran, Iran, in 2014, in mechanical engineering, and the M.Sc. degree from the University of Tehran, Tehran, Iran, in 2017 in mechatronics engineering. His research interests include mechatronics, robotics, control theories, mechanism design, and actuator design.

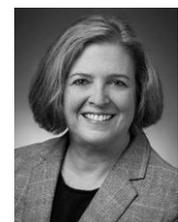

**M. Katherine Banks** is a professor of Civil engineering department and currently Vice Chancellor of Engineering for The Texas A&M University System and Dean of the Texas A&M University College of Engineering. She is an Elected Fellow of the American Society of Civil Engineers, was elected in 2014 to the National Academy of Engineering, and was formerly the Jack and Kay Hockema Professor at Purdue University. Her research interests include applied microbial systems, biofilm processes, wastewater treatment and reuse, and phytoremediation bioremediation. She received her Ph.D. in 1989 from Duke University. At Texas A&M, she helped establish the EnMed program (led by Roderic Pettigrew, Ph.D., M.D.), an innovative engineering medical school option created by Texas A&M University and Houston Methodist Hospital, designed to educate a new kind of physician who will create transformational technology for health care. She received her Bachelor of Science in Engineering from the University of Florida, Master of Science in Engineering from the University of North Carolina, and Doctorate of Philosophy in civil and environmental engineering from Duke University. Dr. Banks is the recipient of the American Society of Civil Engineers Petersen Outstanding Woman of the Year Award, American Society of Civil Engineers Rudolph Hering Medal, Purdue Faculty Scholar Award, Sloan Foundation Mentoring Fellowship and the American Association of University Women Fellowship. On February 13, 2019, she was named to the Board of Directors of Halliburton.